\documentclass[letterpaper, 10 pt, conference]{ieeeconf}  

\IEEEoverridecommandlockouts                              

\title{\LARGE \bf
RoboAfford++: A Generative AI-Enhanced Dataset for Multimodal Affordance Learning in Robotic Manipulation and Navigation
}

\author{Xiaoshuai Hao$^{1}$, Yingbo Tang$^{2,*}$, Lingfeng Zhang$^{3}$, 
Yanbiao Ma$^{4}$, Yunfeng Diao$^{5}$, Ziyu Jia$^{2}$\\
Wenbo Ding$^{3}$, Hangjun Ye$^{1}$, Long Chen$^{1,\dag}$
\thanks{*Corresponding author; \dag Project leader. }
\thanks{$^{1}$Xiaomi EV. E-mail: \{haoxiaoshuai,chenlong37,yehangjun\}@xiaomi.com.}
\thanks{$^{2}$Institute of Automation, Chinese Academy of Sciences. E-mail: tangyingbo2020@ia.ac.cn; jia.ziyu@outlook.com.}%
\thanks{$^{3}$Tsinghua Shenzhen International Graduate School, Tsinghua University. E-mail:zlf25@mails.tsinghua.edu.cn, ding.wenbo@sz.tsinghua.edu.cn.}%
\thanks{$^{4}$Gaoling School of Artificial Intelligence, Renmin University of China. E-mail:ybma1998xidian@gmail.com.}%
\thanks{$^{5}$School of Computer Science and Information Engineering, Hefei University of Technology. E-mail:diaoyunfeng@hfut.edu.cn.}%
}

\usepackage{balance}
\usepackage{multirow}
\usepackage{pifont}
\newcommand{\cmark}{\ding{51}}
\newcommand{\xmark}{\ding{55}}
\usepackage{xspace}
\usepackage{graphicx}
\usepackage{booktabs}
\usepackage[table]{xcolor}
\usepackage{hyperref}

\begin{document}

\maketitle
\thispagestyle{empty}
\pagestyle{empty}

\begin{abstract}

Robotic manipulation and navigation are fundamental capabilities of embodied intelligence, enabling effective robot interactions with the physical world.
In manipulation, predicting precise interactive positions is essential for grasping and placing objects. 
In navigation, find the target and understanding the traversable free space are crucial for safe movement.
Achieving these capabilities requires a cohesive understanding of the environment, including \emph{\textbf{object recognition}} to localize target objects, \emph{\textbf{object affordances}} to identify potential interaction areas and \emph{\textbf{spatial affordances}} to discern optimal areas for both object placement and robot movement. 
While Vision-Language Models (VLMs) excel at high-level task planning and scene understanding, they often struggle to infer actionable positions for physical interaction, such as functional grasping points and permissible placement regions. 
This limitation stems from the lack of fine-grained annotations for object and spatial affordances in their training datasets.
To tackle this challenge, we introduce \textit{\textbf{RoboAfford++}}, a generative AI-enhanced dataset for multimodal affordance learning for both robotic manipulation and navigation. 
Our dataset comprises 869,987 images paired with 2.0 million question answering (QA) annotations, covering three critical tasks: \emph{object affordance recognition} to identify target objects based on attributes and spatial relationships, \emph{object affordance prediction} to pinpoint functional parts for manipulation, and \emph{spatial affordance localization} to identify free space for object placement and robot navigation.
Complementing this dataset, we propose \textbf{\textit{RoboAfford-Eval}}, a comprehensive benchmark for assessing affordance-aware prediction in real-world scenarios, featuring 338 meticulously annotated samples across the same three tasks.
Extensive experimental results reveal the deficiencies of existing VLMs in affordance learning, while fine-tuning on the RoboAfford++ dataset significantly enhances their ability to reason about object and spatial affordances, validating the dataset's effectiveness.
The dataset, benchmark and evaluation code will be made publicly available to facilitate future research. Project website: \url{https://roboafford-dataset.github.io/}.

\end{abstract}

\section{INTRODUCTION}

Enabling robots to physically interact with 3D environments through manipulation and navigation presents a fundamental challenge in embodied AI, requiring advanced perception, reasoning, and adaptability to handle real-world complexities~\cite{li2024foundation,hao2022listen,hao2021matters,wu2025evaluating}. 
Recent advancements in VLMs ~\cite{driess2023palm,hurst2024gpt,gemini2025pro,bai2025qwen2,ji2025robobrain,team2025gemini,team2025robobrain,azzolini2025cosmos} have significantly improved robots' perceptual and reasoning capabilities, facilitating progress in scene understanding and task planning. These models have shown considerable potential in various embodied tasks, including vision-language navigation~\cite{zhou2024navgpt,zhang2025mapnav,chen2025affordances,zhang2024trihelper,zhang2025multi,zhang2025nava,xiao2025team,zhang2025team,liu2025toponav,gong2025stairway}, embodied manipulation~\cite{zawalski2024robotic,xu2025a0,tang2025affordgrasp,hao2025tla,zhang2025vtla,tang2025roboafford,team2025robobrain,zhao2025training}, and embodied reasoning~\cite{team2025gemini,zhang2025embodied,tan2025reason}.
However, despite these advancements in high-level planning and semantic understanding, existing VLMs still encounter significant challenges in achieving fine-grained interaction understanding for both manipulation and navigation scenarios.

\begin{figure*}[!ht]
\centering
  \includegraphics[width=0.94\textwidth]{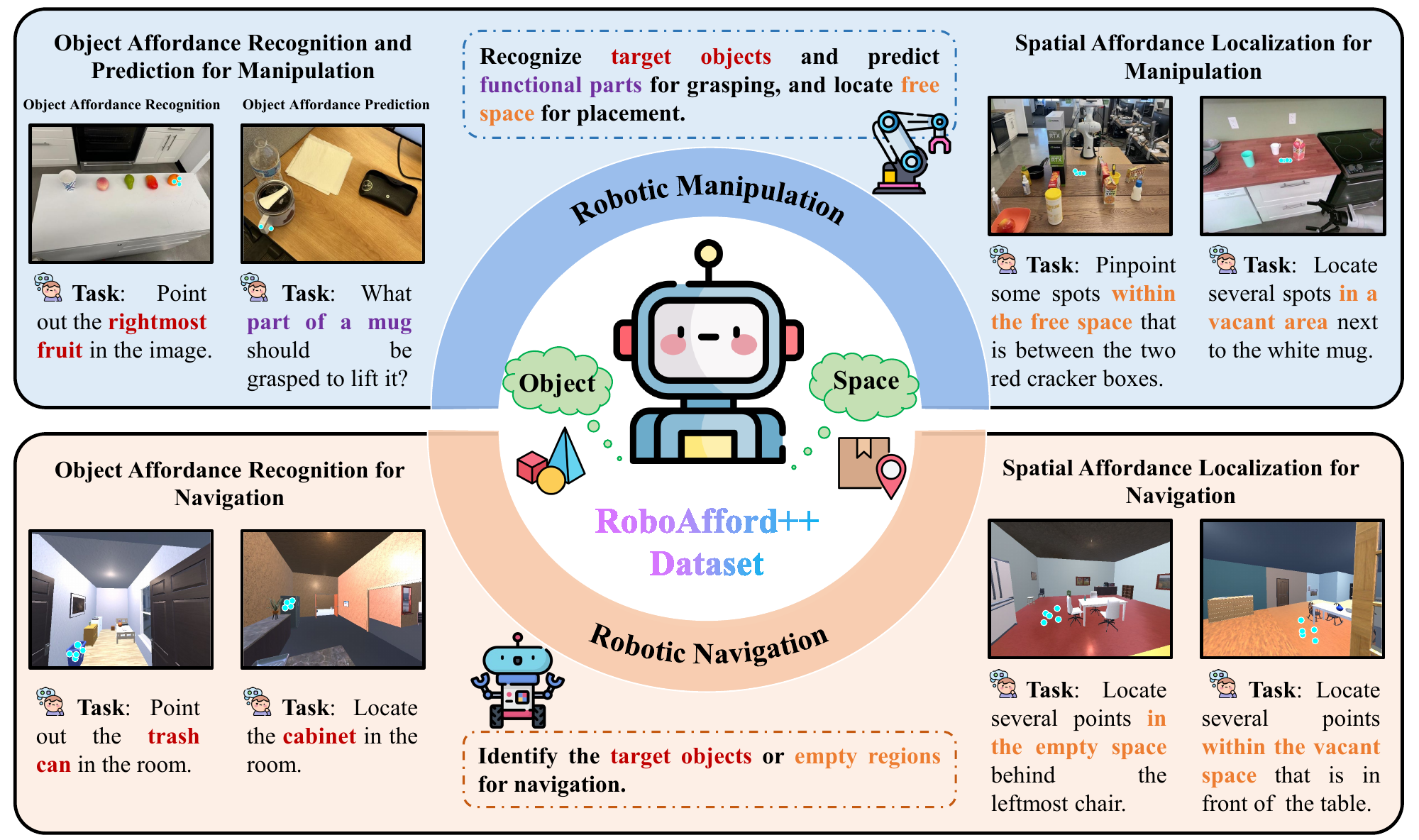}
  \vspace{-4pt}
  \caption{\textbf{Overview of the RoboAfford++ Dataset.} It encompasses three key capabilities: object affordance recognition, object affordance prediction, and spatial affordance localization for both manipulation and navigation tasks.}
  \label{figure1}
  \vspace{-4pt}
\end{figure*}

In robotic manipulation and navigation, VLMs are required to ground target objects or their parts from language instructions while identifying suitable regions for placement or movement. This involves three core capabilities: (1) \textbf{\textit{Object Affordance Recognition}}: identifying objects based on attributes such as category, color, size, and spatial relations; (2) \textbf{\textit{Object Affordance Prediction}}: localizing functional parts of objects to support specific actions, such as the handle of a teapot for grasping; and (3) \textbf{\textit{Spatial Affordance Localization}}: detecting vacant areas in the scene for object placement and robot navigation, such as empty space in shelves for storing items.
Despite recent progress~\cite{yuan2024robopoint,song2024robospatial,tang2025roboafford}, relatively few works have achieved a holistic integration of object and spatial affordances that adequately supports both manipulation and navigation tasks. While some VLMs~\cite{yuan2024robopoint,song2024robospatial} can coarsely localize objects or assess spatial compatibility through coordinates, they often fail to perform fine-grained localization at the part level, such as identifying functional components of objects. Furthermore, the models are specifically trained for manipulation tasks, resulting in a narrow focus that limits their generalization across diverse manipulation and navigation scenarios in real world.

To address these challenges, we present \textbf{\textit{RoboAfford++}}, a generative AI-enhanced large-scale dataset with \textit{dense, affordance-aware annotations} for robotic manipulation and navigation. It contains 869,987 images and 2.0 million QA pairs, unifying object and spatial affordances to support interaction-centric learning across both domains..
As shown in Fig.~\ref{figure1}, RoboAfford++ aims to equip VLMs with affordance reasoning capabilities essential for embodied interaction. The dataset enables models to ground target objects and predict part-level affordances for interaction, and identify spatial affordances for both placement and navigation. 
By employing generative AI techniques for data augmentation and annotation, RoboAfford++ specifically addresses the challenge of translating high-level instructions into actionable positions for manipulation and feasible areas for navigation.
Additionally, we introduce \textbf{\textit{RoboAfford-Eval}}, a rigorously annotated benchmark featuring 338 evaluation questions. RoboAfford-Eval encompasses three tasks: (1) object affordance recognition to assess grounding ability, (2) object affordance prediction to evaluate functional part understanding, and (3) spatial affordance localization to test free space detection. 
Extensive experiments demonstrate that fine-tuning on the RoboAfford++ dataset significantly enhances the model's affordance reasoning capabilities for both manipulation and navigation tasks. Our contributions are summarized as follows:

\begin{table*}[h]
\centering
  \caption{Comparison of existing affordance datasets. \textbf{Obj-Aff}: Object Affordance. \textbf{Spa-Aff}: Spatial Affordance.}
  \label{table:affordance_dataset}
  \resizebox{0.92\textwidth}{!}{
  \begin{tabular}{lccccccccc}
    \toprule
    \textbf{Dataset} & \textbf{Domain} & \textbf{Format} & \textbf{Obj-Aff}  & \textbf{Spa-Aff}  & \textbf{\#Spatial Relations} & \textbf{\#Images} & \textbf{\#QAs} \\
    \midrule
     UMD~\cite{myers2015affordance} & Tabletop & 2D Mask & \cmark & \xmark & - & 30.0K & - \\
     IIT-AFF~\cite{nguyen2017object} & Generic & 2D Mask & \cmark & \xmark & - & 8.8K & - \\
     PAD~\cite{luo2021one} & Generic & 2D Mask & \cmark & \xmark & - & 4.0K & - \\
     PADv2~\cite{zhai2022one} & Generic & 2D Mask & \cmark & \xmark & - & 30.0K & - \\
     AGD-20K~\cite{luo2022learning} & Generic & 2D Heatmap & \cmark & \xmark & - & 26.1K & - \\
     3DOI~\cite{qian2023understanding} & Generic, indoor & 2D Mask, point & \cmark & \xmark & - & 10.0K & - \\
     RoboPoint~\cite{yuan2024robopoint} & Generic, tabletop & 2D Point & \xmark & \cmark  & 14 & - & 1.4M \\
     RoboSpatial~\cite{song2024robospatial} & Indoor, tabletop & 2D Point & \xmark & \cmark  & 6 & 1.0M & 3.0M \\
     \rowcolor{blue!8} \textbf{RoboAfford++ (Ours)} & \textbf{Generic, indoor, tabletop} & \textbf{2D Point} & \cmark & \cmark & \textbf{20} & \textbf{870.0K} & \textbf{2.0M} \\
    \bottomrule
  \end{tabular}
  }
  \vspace{-0.5em}
\end{table*}

\begin{itemize}
\item 
We introduce \textbf{\textit{RoboAfford++}}, a novel generative AI-enhanced dataset that unifies object and spatial affordances for both robotic manipulation and navigation. It comprises 2.0 million question answering pairs with precise 2D point annotations, focusing on three key capabilities: object affordance recognition, object affordance prediction, and spatial affordance localization. This comprehensive dataset fills the critical gap in fine-grained affordance annotations necessary for precise physical interactions.

\item 
We present the \textbf{\textit{RoboAfford-Eval}} benchmark, featuring 338 manually annotated questions designed to systematically evaluate VLMs on three core tasks: object affordance recognition, object affordance prediction, and spatial affordance localization. This benchmark establishes criteria to assess both object-level and scene-level affordance understanding.

\item 
Extensive experiments reveal the limitations of existing VLMs in affordance learning, while fine-tuning on the RoboAfford++ dataset significantly enhances their affordance prediction for both manipulation and navigation scenarios, confirming its effectiveness and practical value.
\end{itemize}

\section{RELATED WORK}

\textbf{Datasets for Affordance Learning}
Affordance learning is a key research area in computer vision and robotics, which focuses on understanding object and environment properties for interaction. Prior works~\cite{li2023locate,qian2024affordancellm,jiang2025affordancesam} primarily aimed to identify functional object parts enabling specific interactions through semantic segmentation, using pixel-wise annotated datasets like UMD~\cite{myers2015affordance}, IIT-AFF~\cite{nguyen2017object}, PAD~\cite{luo2021one}, PADv2~\cite{zhai2022one}, AGD-20K~\cite{luo2022learning}, and 3DOI~\cite{qian2023understanding}.
Another approach utilizes videos to transfer learned affordances to target images via interactive heatmap predictions~\cite{bahl2023affordances,chen2023affordance}. Some works also explored 3D affordance datasets~\cite{deng20213d,li2024laso,delitzas2024scenefun3d}, modeling affordances as grasping contact maps and scene-aware motion synthesis.
However, existing datasets often focus on either object-centric or scene-centric annotations, rarely unifying both, as shown in Tab.~\ref{table:affordance_dataset}. This paper proposes the RoboAfford++ dataset to bridge this gap. It enables the model to predict interactive points by unifying object affordances (e.g., target objects and functional parts) and spatial affordances (e.g., vacant areas) into a cohesive framework for real-world scene understanding.

\textbf{Spatial Reasoning with VLMs}
Spatial reasoning is crucial for robots to interact effectively with the physical world. Building on 2D VLMs, several works~\cite{chen2024spatialvlm,cheng2024spatialrgpt,yuan2024robopoint,cai2024spatialbot,song2024robospatial,liu2025spatialcot,zhang2025video,wu2025spatialscore} have extracted 3D spatial information from 2D images to generate large-scale, spatially-related question answering pairs for fine-tuning. For example, SpatialVLM~\cite{chen2024spatialvlm} converts 2D images into object-centric 3D point clouds using metric depth estimation, synthesizing VQA data with 3D spatial reasoning supervision. SpatialRGPT~\cite{cheng2024spatialrgpt} enhances region-level spatial reasoning through region proposals and 3D scene graph construction. RoboPoint~\cite{yuan2024robopoint} introduces a synthetic dataset for object and free space reference, predicting points on objects or vacant regions that satisfy certain spatial relations. Recent studies like SpatialBot~\cite{cai2024spatialbot} and RoboSpatial~\cite{song2024robospatial} utilize both RGB and depth images to improve spatial understanding. Additionally, SpatialCoT~\cite{liu2025spatialcot} leverages language models to generate rationales for complex spatial reasoning. 
Despite these advancements, existing VLMs exhibit limited spatial reasoning for real-world interactions. Their insufficient coverage of object functionality and spatial relationships restricts accurate affordance predictions. 
To address this, we explore enhancing VLMs' capabilities through large-scale, generative AI-enhanced data specifically designed for object and spatial affordance learning.

\section{RoboAfford++ Dataset and Benchmark Construction}
\subsection{Data Collection and Filtering}
The RoboAfford++ dataset integrates annotations from six data sources, as shown in Tab.~\ref{table:dataset_source}. 
By combining real-world data from the internet with synthetic data generated in simulation, we create a diverse dataset for modeling object and spatial interactions. This dataset consists of three main components: Object Affordance Recognition, Object Affordance Prediction, and Spatial Affordance Localization.

\begin{table}[t]
\centering
\caption{Data source information for RoboAfford++ dataset.}
\resizebox{0.46\textwidth}{!}{%
\begin{tabular}{lcc}
\toprule
\textbf{Data Source} & \textbf{\#Selected Images} & \textbf{\#Generated QA Pairs} \\
\midrule
LVIS~\cite{gupta2019lvis} & 152,152 & 513K \\
Pixmo-Points~\cite{deitke2024molmo} & 63,907 & 190K \\
PACO-LVIS~\cite{ramanathan2023paco} & 45,790 & 561K \\
Object Reference~\cite{yuan2024robopoint} & 287,956 & 347K \\
Region Reference~\cite{yuan2024robopoint} & 270,182 & 320K \\
NaviAfford (AI Generated) & 50,000 & 100K \\
\bottomrule
\end{tabular}%
}
\label{table:dataset_source}
\vspace{-1.0em}
\end{table}

\begin{figure*}[!ht]
\centering
  \includegraphics[width=0.9\textwidth]{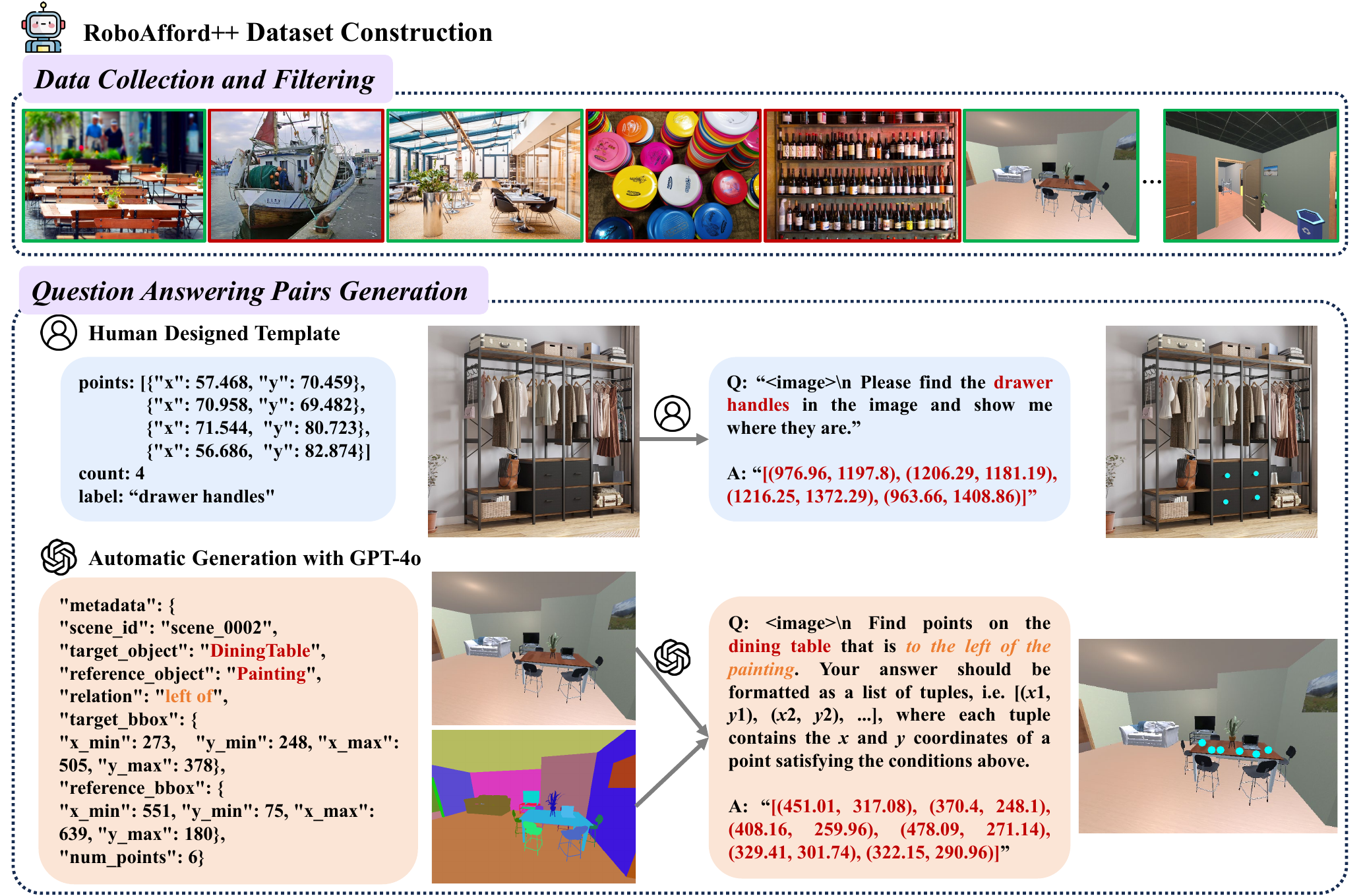}
  \caption{Pipeline for constructing the RoboAfford++ dataset. We first discard the image with densely repeated objects, and then generate question answering pairs using human designed template or GPT-4o~\cite{hurst2024gpt}.}
  \label{figure_data_construction}
\end{figure*}

\textbf{Object Affordance Recognition}  
To equip VLMs with general object recognition knowledge, we utilize LVIS~\cite{gupta2019lvis}, which offers 2.2M high-quality instance segmentation masks across over 1,000 categories. We convert 152K images from LVIS~\cite{gupta2019lvis} into an object detection format with bounding box coordinates \((x_1,y_1,x_2,y_2)\) to establish visual grounding. Additionally, we leverage the Pixmo-Points dataset~\cite{deitke2024molmo}, containing 2.3M point annotations from 223K images, along with 288K synthetic images from RoboPoint~\cite{yuan2024robopoint} for object reference learning. To address the issue of densely repeated object instances in Pixmo-Points~\cite{deitke2024molmo}, we implement a two-step filtering process: first, we discard annotations with more than 10 point labels for training simplicity, and second, we use GPT-4o~\cite{hurst2024gpt} to retain only relevant indoor objects (\textit{e.g.}, furniture, kitchenware), resulting in 63,907 images suitable for object pointing. 

Designed for navigation task, we constructed the NaviAfford dataset in the AI2Thor simulator~\cite{kolve2017ai2}, consisting of 50,000 egocentric images captured across 200 indoor environments. Our data collection process begins by randomly selecting accessible positions while excluding regions with sufficient clearance (more than 1.5 meters). From each location, we gather RGB images along with instance segmentation masks, sampling a diverse range of viewing angles (0–360° random rotation, and -15° to 15° horizontal rotation). Each image is accompanied by metadata such as bounding boxes of visible objects, their 3D distances, and 2D coordinates. We then derive spatial relationship annotations by detecting object pairs that satisfy predefined proximity criteria (\textit{e.g.}, a horizontal separation of more than 20 pixels). For every valid relation, we produce 4 to 8 pointing locations within the target object’s bounding box, forming instructions phrased as ``locate several points on \texttt{<target object>} \texttt{<relation>} \texttt{<reference object>}.'' In total, the NaviAfford dataset provides 50,000 affordance samples tailored for navigation training.

\textbf{Object Affordance Prediction}  
For object affordance prediction, we utilize the PACO-LVIS dataset~\cite{ramanathan2023paco} to provide part-level annotations for reasoning. We extract bounding box and part segmentation masks from 45,790 images across 75 object categories and 200 part categories, transforming these into ground truth labels for object affordance. This structured data enables precise predictions of how objects can be interacted with based on their affordances.

\textbf{Spatial Affordance Localization}  
For spatial affordance localization, we source region reference data from RoboPoint~\cite{yuan2024robopoint}, utilizing 270K images across 8K instances and 262 categories. Each image includes one or two colored bounding boxes to indicate objects, with ground truths formulated as series of points \([(x_1,y_1),(x_2,y_2),...]\) for free space referencing. We convert normalized coordinates to absolute positions and sample ground truths to a maximum of ten points per answer, enhancing model optimization for spatial tasks.

\subsection{Question Answering Pairs Generation}
We create specialized question-answer generation pipelines for each task in RoboAfford++, as shown in Fig.~\ref{figure_data_construction}. By transforming the collected data into affordance-aware QAs, we enhance the engagement of VLMs with the dataset, allowing them to learn and reason about spatial relationships between objects and their affordances.

\textbullet~ For object affordance recognition, we use GPT-4o~\cite{hurst2024gpt} to analyze scenes and filter out irrelevant outdoor images. We create templates for generating questions and answers, such as ``Point to all occurrences of \texttt{<label>} in the image'' and ``Can you see any \texttt{<label>} in the image? Point to them,'' where ``\texttt{<label>}'' refers to ground truths from Pixmo-Points~\cite{deitke2024molmo}. We design 28 templates and randomly select one for each object pointing question.

\textbullet~ For object affordance prediction, we generate QAs by prompting GPT-4o~\cite{hurst2024gpt} with images, object and part categories, and ground truth bounding boxes. Questions for whole objects focus on functionality without naming the object (\textit{e.g.}, ``What appliance can be used to heat food quickly?'' for a microwave). For object parts, we ask for part identification (\textit{e.g.}, ``Which part of a knife should be held to cut safely?'' for the knife's handle). Ground truth answers include two formats: (1) bounding boxes for the target object or part, and (2) points sampled from the ground truth segmentation mask. This dual representation ensures accurate part grounding and enhances fine-grained object affordance prediction.

\textbullet~ For spatial affordance localization, we generate QAs by modifying annotations from RoboPoint~\cite{yuan2024robopoint}. Specifically, we convert normalized coordinates to absolute positions for accurate real-world object scales and spatial relationships. Ground-truth points are resampled to a maximum of ten per question, and instructions are adjusted for answer format consistency. This method preserves the spatial relationships defined in RoboPoint~\cite{yuan2024robopoint} while integrating them into our unified affordance framework.

\begin{figure*}[!ht]
\centering
  \includegraphics[width=0.96\textwidth]{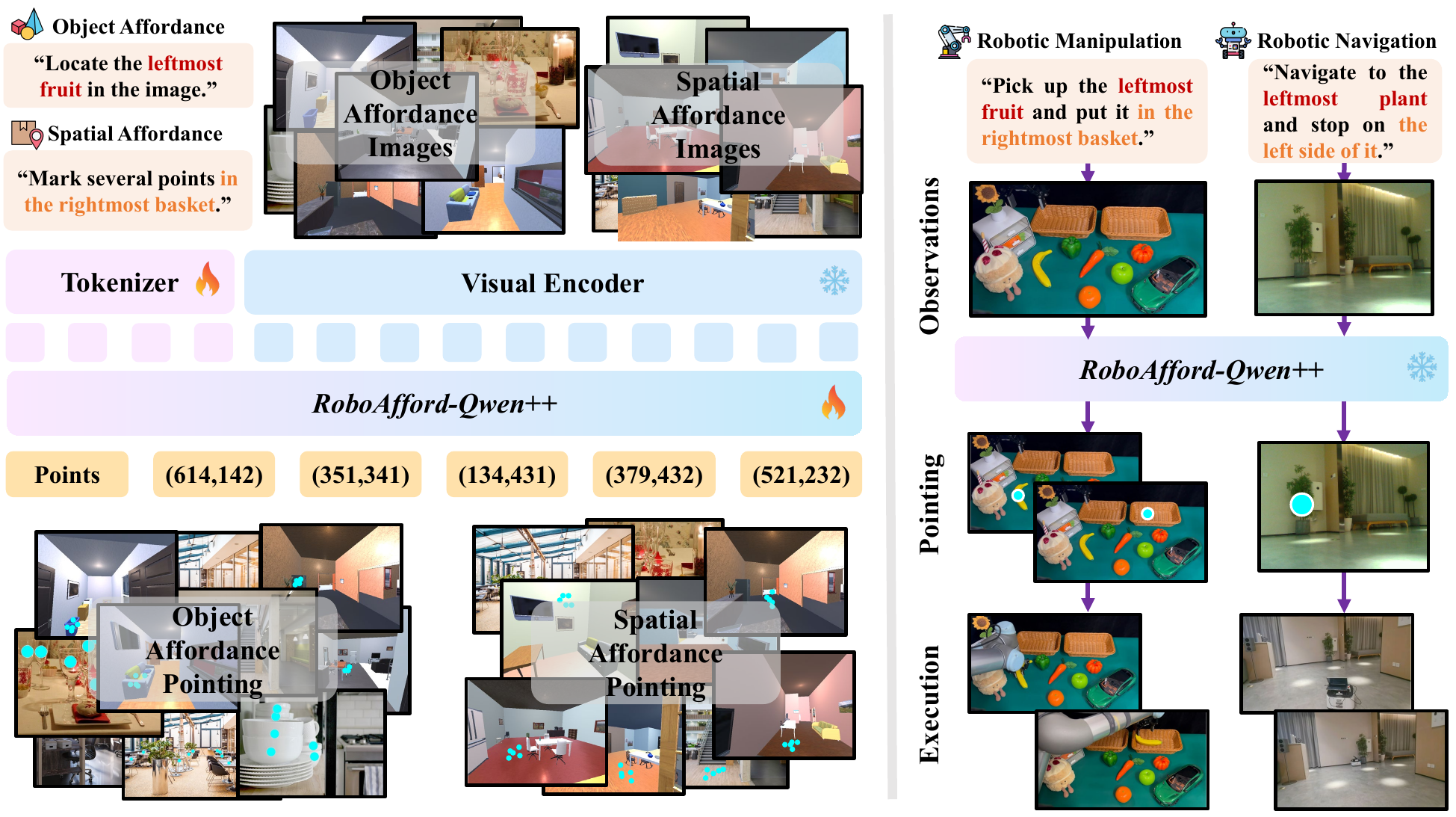}
  \caption{Framework of RoboAfford-Qwen++. We fine-tune the model on the RoboAfford++ dataset to enhance object and spatial affordance capabilities. For downstream robotic  manipulation and navigation tasks, we integrate depth images to convert 2D points representing affordances into 3D coordinates, which are then used as target positions for robotic execution.}
  \label{figure_framework}
  \vspace{-4pt}
\end{figure*}

\subsection{RoboAfford-Eval Benchmark}
To evaluate object affordance recognition and prediction, we manually annotated 114 questions for recognition and 124 for prediction using images from the Where2Place~\cite{yuan2024robopoint} dataset. For spatial affordances, we maintained the original 100 questions but converted the annotated affordance points from normalized to absolute coordinate representations. Ground-truth for each question consists of one or more human-annotated polygon masks corresponding to the parts or instances in the answers.

For each predicted point, we check if it falls within the ground- truth masks. The accuracy for a question is the ratio of correctly located points to total predicted points, with overall accuracy being the average across all questions. To enforce stricter criteria, we penalize points outside the image boundaries, encouraging the model to better learn absolute interactive positions.

\section{Methodology}
\textbf{Framework}
Using the RoboAfford++ dataset, we fine-tune a model named RoboAfford-Qwen++. This model employs a multimodal architecture comprising a vision encoder, an MLP projector, a language tokenizer, and the Qwen2.5 Large Language Model (LLM) following~\cite{bai2025qwen2}, as illustrated in Fig.~\ref{figure_framework}. The vision encoder extracts visual features from input images, which are transformed into the same embedding space as the language tokens through the MLP projector. These visual embeddings are concatenated with the embedded textual instructions and input to the LLM for joint reasoning across modalities.

\textbf{Instruction Fine-tuning}
We employ a multi-stage training strategy based on the LLaVA-1.5 instruction tuning framework~\cite{liu2023visual}, consisting of two phases: General Localization Learning and Object-Spatial Affordance Enhancement. The first phase uses the LVIS~\cite{gupta2019lvis} and Pixmo-Points~\cite{deitke2024molmo} datasets, totaling 216K images and 703K QA pairs, to enhance basic object affordance recognition. The second phase incorporates Object Reference~\cite{yuan2024robopoint}, NaviAfford, PACO-LVIS~\cite{ramanathan2023paco}, and Region Reference~\cite{yuan2024robopoint} data, amounting to 654K images and 1.33M QA pairs, to optimize object affordance prediction and spatial affordance localization. This multi-stage approach allows the model to progressively develop hierarchical affordance reasoning, evolving from basic recognition to advanced prediction tasks.

\textbf{Real-world Robotic Manipulation and Navigation}
Fig.~\ref{figure_framework} shows that the fine-tuned RoboAfford-Qwen++ can be effectively applied to downstream robotic manipulation and navigation. For the task ``Pick up the leftmost fruit and put it in the rightmost basket,'' RoboAfford-Qwen++ predicts the object's affordance for the specified fruit and the spatial affordance of the basket for feasible placement. The predicted 2D affordance points are then used as prompts to segment the target object by~\cite{ravi2024sam} or transformed into 3D coordinates using a depth map, where the grasping poses are generated by~\cite{fang2023anygrasp} for robotic execution.

\begin{table*}[t]
\centering
\caption{Comparison results of various VLMs on RoboAfford-Eval benchmark.}
\label{table:model_performance}
\resizebox{0.98\textwidth}{!}{
\begin{tabular}{c|l|c|ccc|c}
\toprule
\multirow{2}{*}{\textbf{Type}} & \multirow{2}{*}{\textbf{Models}} & \multirow{2}{*}{\textbf{Parameters}} & \textbf{Object Affordance} & \textbf{Object Affordance} & \textbf{Spatial Affordance} & \multirow{2}{*}{\textbf{Average$\uparrow$}} \\
&  &  & \textbf{Recognition$\uparrow$} & \textbf{Prediction$\uparrow$} & \textbf{Localization$\uparrow$} & \\
\midrule
\multirow{4}{*}{\parbox{1cm}{Closed-source Models}}
& \cellcolor{red!5} GPT-4o~\cite{hurst2024gpt} & \cellcolor{red!5} - & \cellcolor{red!5} 21.2 & \cellcolor{red!5} 15.9 & \cellcolor{red!5} 25.4 & \cellcolor{red!5} 20.5 \\
& \cellcolor{red!5} Claude-3.5-Sonnet~\cite{anthropic2024claude} & \cellcolor{red!5} - & \cellcolor{red!5} 20.4 & \cellcolor{red!5} 13.1 & \cellcolor{red!5} 22.6 & \cellcolor{red!5} 18.4 \\
& \cellcolor{red!5} Gemini-2.5-Flash~\cite{gemini2025flash} & \cellcolor{red!5} - & \cellcolor{red!5} 20.4 & \cellcolor{red!5} 21.7 & \cellcolor{red!5} 29.4 & \cellcolor{red!5} 23.5 \\
& \cellcolor{red!5} Gemini-2.5-Pro~\cite{gemini2025pro} & \cellcolor{red!5} - & \cellcolor{red!5} 17.0 & \cellcolor{red!5} 14.5 & \cellcolor{red!5} 41.8 & \cellcolor{red!5} 23.4 \\
\midrule
\multirow{8}{*}{\parbox{1cm}{Open-source Models}}
& \cellcolor{yellow!8} Qwen2.5-VL~\cite{bai2025qwen2} & \cellcolor{yellow!8} 3B & \cellcolor{yellow!8} 7.1 & \cellcolor{yellow!8} 2.0 & \cellcolor{yellow!8} 12.5 & \cellcolor{yellow!8} 6.8 \\
& \cellcolor{yellow!8} Molmo~\cite{deitke2024molmo} & \cellcolor{yellow!8} 7B & \cellcolor{yellow!8} 5.7 & \cellcolor{yellow!8} 4.7 & \cellcolor{yellow!8} 4.7 & \cellcolor{yellow!8} 5.0 \\
& \cellcolor{yellow!8} Qwen2-VL~\cite{wang2024qwen2} & \cellcolor{yellow!8} 7B & \cellcolor{yellow!8} 15.6 & \cellcolor{yellow!8} 10.5 & \cellcolor{yellow!8} 14.3 & \cellcolor{yellow!8} 13.4 \\
& \cellcolor{yellow!8} LLaVA-Next~\cite{liu2024improved} & \cellcolor{yellow!8} 8B & \cellcolor{yellow!8} 2.9 & \cellcolor{yellow!8} 0.8 & \cellcolor{yellow!8} 0.6 & \cellcolor{yellow!8} 1.4 \\
& \cellcolor{yellow!8} SpaceMantis~\cite{chen2024spatialvlm} & \cellcolor{yellow!8} 8B & \cellcolor{yellow!8} 3.6 & \cellcolor{yellow!8} 4.8 & \cellcolor{yellow!8} 12.0 & \cellcolor{yellow!8} 6.5 \\
& \cellcolor{yellow!8} RoboPoint~\cite{yuan2024robopoint} & \cellcolor{yellow!8} 13B & \cellcolor{yellow!8} 55.7 & \cellcolor{yellow!8} 35.0 & \cellcolor{yellow!8} 44.2 & \cellcolor{yellow!8} 44.7 \\
\cmidrule(lr){2-7}
& \cellcolor{blue!3} Qwen2.5-VL~\cite{bai2025qwen2} (Baseline) & \cellcolor{blue!3} 7B & \cellcolor{blue!3} 19.5 & \cellcolor{blue!3} 8.3 & \cellcolor{blue!3} 21.9 & \cellcolor{blue!3} 16.1 \\
 & \cellcolor{blue!6} RoboAfford-Qwen~\cite{tang2025roboafford} & \cellcolor{blue!6} 7B
 & \cellcolor{blue!6} 66.1 (+46.6$\uparrow$) & \cellcolor{blue!6} 54.3 (+46.0$\uparrow$) & \cellcolor{blue!6} \textbf{57.9 (+36.0$\uparrow$)} & \cellcolor{blue!6} 59.3 (+43.2$\uparrow$) \\
  & \cellcolor{blue!8} \textbf{RoboAfford-Qwen++ (Ours)} & \cellcolor{blue!8} \textbf{7B}
 & \cellcolor{blue!8} \textbf{70.5 (+51.0$\uparrow$)} & \cellcolor{blue!8} \textbf{63.1 (+54.8$\uparrow$)} & \cellcolor{blue!8} 55.8 (+33.9$\uparrow$) & \cellcolor{blue!8} \textbf{63.4 (+47.3$\uparrow$)} \\
\bottomrule
\end{tabular}}
\end{table*}

\section{Experiments}
\subsection{Experimental Setup}
\textbf{Implementation Details.} Our RoboAfford-Qwen++ model is initialized with the pre-trained Qwen2.5-VL-7B-Instruct~\cite{qwen2.5-VL} weights and undergoes full-parameter supervised fine-tuning as described in~\cite{zheng2024llamafactory}. The experiments are conducted on eight H100 GPUs using AdamW~\cite{adamw} as the optimizer, with a learning rate of \(10^{-5}\) and a training duration of 1 epoch. Each device processes a batch size of 4, with gradient accumulation set to 2 steps.

\textbf{Baseline Models.} We evaluate a range of state-of-the-art VLMs, including both closed-source and open-source models, using the proposed RoboAfford-Eval benchmark. Closed-source models include GPT-4o~\cite{hurst2024gpt}, Claude-3.5-Sonnet~\cite{anthropic2024claude}, Gemini-2.5-Flash~\cite{gemini2025flash}, and Gemini-2.5-Pro~\cite{gemini2025pro}. Open-source models encompass general-purpose VLMs such as LLaVA-Next~\cite{liu2024improved}, Molmo~\cite{deitke2024molmo}, Qwen2-VL~\cite{wang2024qwen2}, and Qwen2.5-VL~\cite{bai2025qwen2}. We also evaluate open-source VLMs with spatial awareness, including SpaceMantis (a community implementation of SpatialVLM~\cite{chen2024spatialvlm}), RoboPoint~\cite{yuan2024robopoint}, and RoboAfford-Qwen~\cite{tang2025roboafford}.

\textbf{Evaluation Metrics.} We evaluate the proposed RoboAfford-Eval benchmark across three tasks: object affordance recognition, object affordance prediction, and spatial affordance localization. The evaluation metric used is accuracy (Acc), defined as the ratio of correctly located predicted points within the ground truth masks to the total number of predicted points. For real-world manipulation and navigation, we adopt the success rate (SR), defined as the ratio of successful executions to the total attempts.

\subsection{Comparison of State-of-the-Art Models}
Tab.~\ref{table:model_performance} presents a quantitative comparison of various baseline models on the RoboAfford-Eval benchmark. The zero-shot generalization ability of generic VLMs is limited across all object-spatial affordance settings, with top competitors like Gemini-2.5-Flash~\cite{gemini2025flash} and Gemini-2.5-Pro~\cite{gemini2025pro} achieving only 23.5 and 23.4 average accuracy, respectively. In contrast, specialized VLM with spatial reasoning capabilities such as RoboPoint~\cite{yuan2024robopoint} shows improved performance, resulting in an average accuracy of 44.7. This underscores the critical role of spatial reasoning in affordance tasks.

Our model RoboAfford-Qwen++, which is based on Qwen2.5-VL-7B~\cite{bai2025qwen2} and fine-tuned on RoboAfford++ dataset, generally outperforms all baseline models, achieving accuracies of 70.5, 63.1, and 55.8 on three tasks, resulting in an average accuracy of 63.4. Compared to baseline model Qwen2.5-VL-7B~\cite{bai2025qwen2}, our RoboAfford-Qwen++ improves the accuracies by 51.0, 54.8, 33.9 on three tasks, respectively. It shows an improvement of 18.7 in average accuracy over the specialized VLM RoboPoint~\cite{yuan2024robopoint}, demonstrating the effectiveness of the RoboAfford++ dataset in enhancing object-spatial affordance understanding.

\begin{figure}[h!]
\centering
  \includegraphics[width=0.46\textwidth]{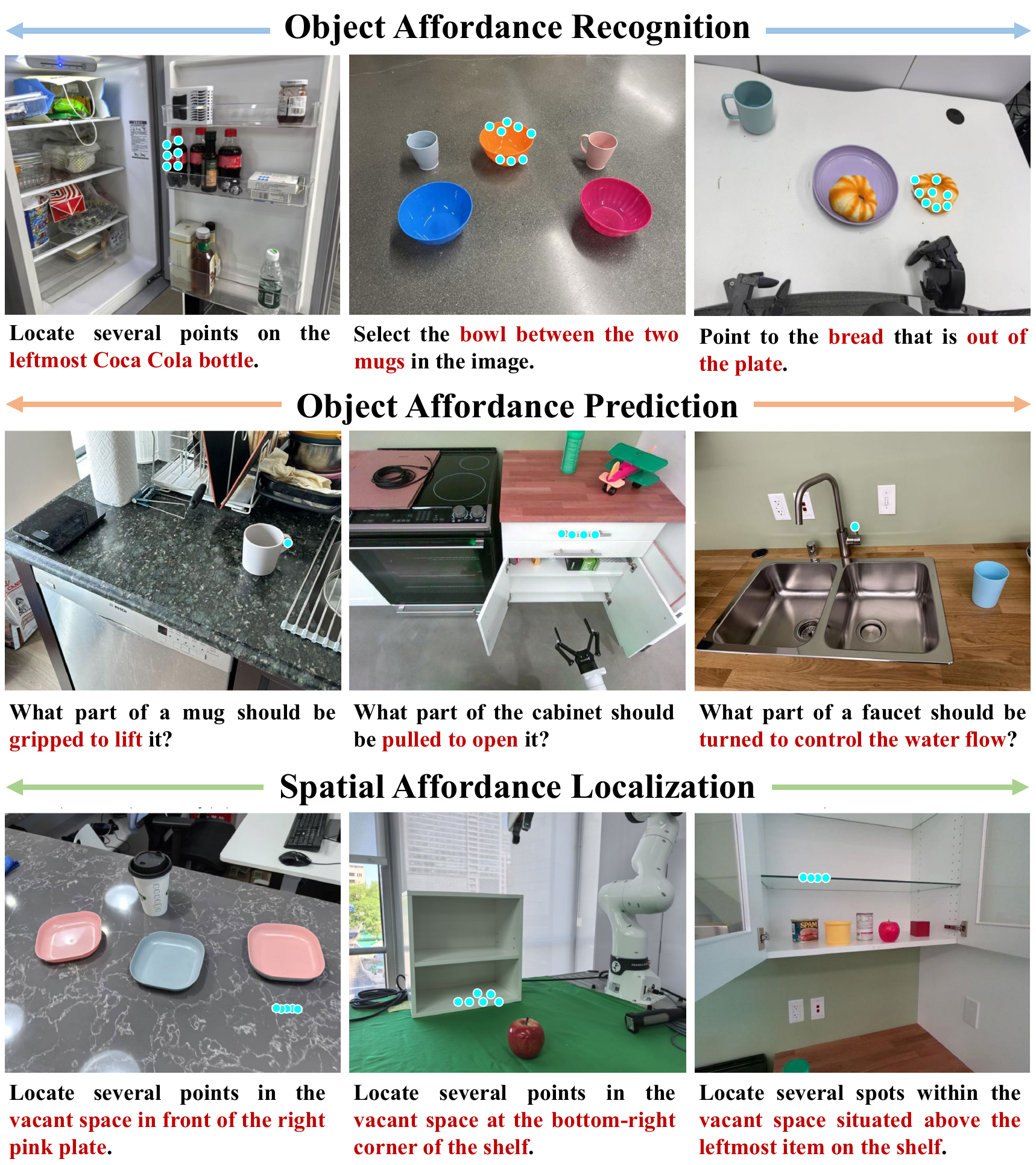}
  \caption{Qualitative results of RoboAfford-Qwen++, where cyan points indicate the object and spatial affordances.}
  \label{figure_results}
\vspace{-1.0em}
\end{figure}

\subsection{Qualitative Results}
To intuitively demonstrate the capabilities of RoboAfford-Qwen++, we present qualitative results in Fig.~\ref{figure_results}. The visualizations illustrate how RoboAfford-Qwen++ adapts to various real-world scenarios, perspectives, and tasks. The model effectively identifies object parts and aligns spatial semantics with task instructions, showcasing its understanding from object-level affordances to scene-level spatial reasoning.

\begin{figure*}[!t]
\centering
  \includegraphics[width=0.98\textwidth]{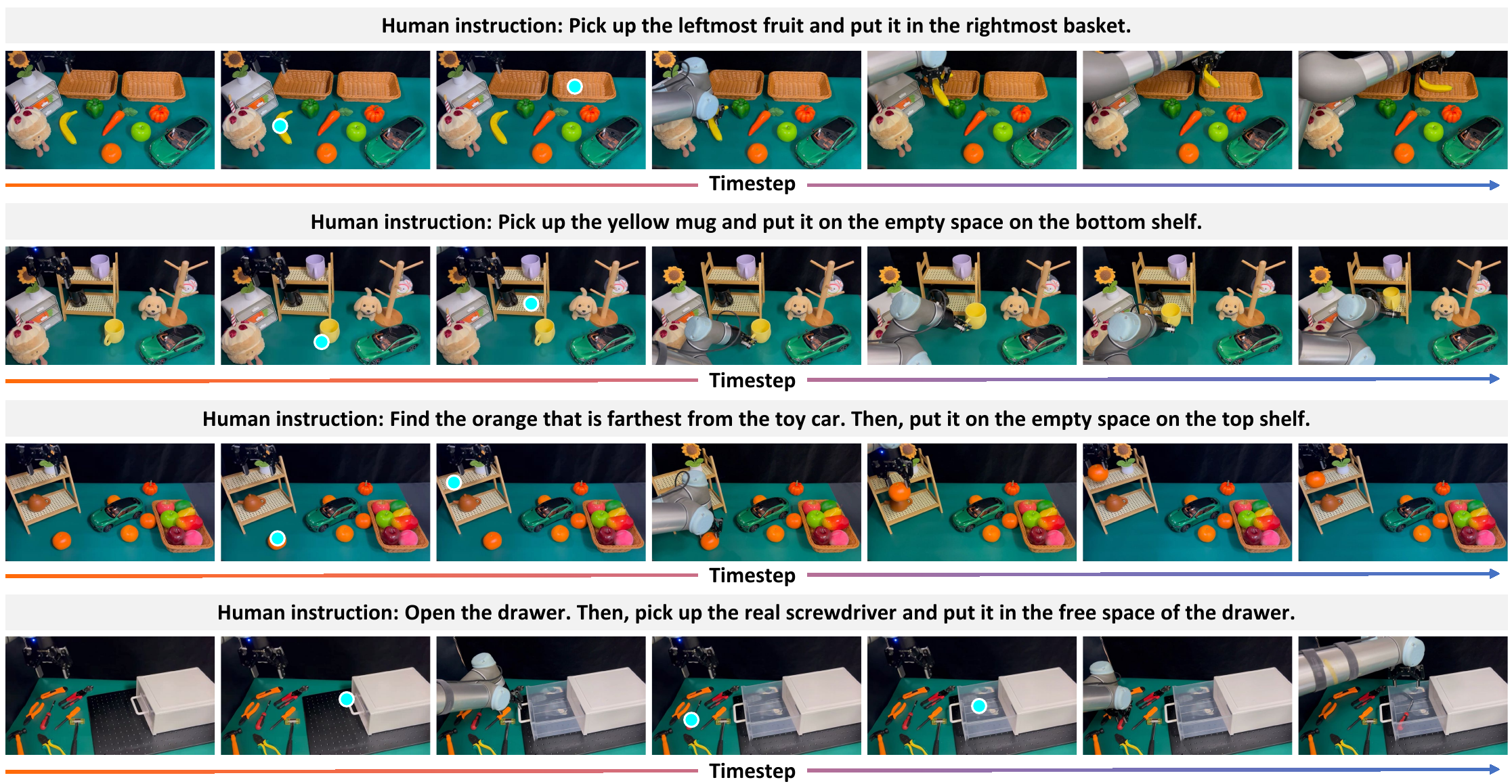}
  \caption{Results of deploying RoboAfford-Qwen++ model to downstream robotic manipulation tasks.}
  \label{figure_manipulation}
\end{figure*}

\begin{table*}[!t]
\centering
\caption{Real-world manipulation comparison of different methods.}
\label{table:realworld_manipulation}
\resizebox{0.96\textwidth}{!}{
\begin{tabular}{l|ccccccc|c}
\toprule
\textbf{Model} & 
\multicolumn{1}{c}{\textbf{Apple}} & 
\multicolumn{1}{c}{\textbf{Mug}} & 
\multicolumn{1}{c}{\textbf{Drawer}} &
\multicolumn{1}{c}{\textbf{in Bowl}} & 
\multicolumn{1}{c}{\textbf{on Plate}} & 
\multicolumn{1}{c}{\textbf{in Cabinet (top)}} &
\multicolumn{1}{c|}{\textbf{in Cabinet (bottom)}} &
\multicolumn{1}{c}{\textbf{Avg. SR$\uparrow$}} \\
\midrule
GPT-4o~\cite{hurst2024gpt}  & 2/10 & 1/10 & 0/10 & 2/10 & 2/10 & 1/10 & 0/10 & 11.4\% \\
Qwen2.5-VL-7B~\cite{bai2025qwen2}   & 1/10 & 0/10 & 0/10 & 2/10 & 1/10 & 0/10 & 0/10 & 5.7\% \\
RoboPoint~\cite{yuan2024robopoint}   & 5/10 & 2/10 & 1/10 & 6/10 & 6/10 & 2/10 & 3/10 & 35.7\% \\
\midrule
\rowcolor{blue!5}
\textbf{RoboAfford-Qwen++ (Ours)}    & \textbf{7/10} & \textbf{5/10} & \textbf{4/10} & \textbf{8/10} & \textbf{8/10} & \textbf{6/10} & \textbf{5/10} & \textbf{61.4\% (+25.7\%$\uparrow$)}\\
\bottomrule
\end{tabular}}
\end{table*}

\subsection{RoboAfford-Qwen++ for Downstream Robotic Tasks}
\textbf{Robotic Manipulation} We rigorously evaluate the RoboAfford-Qwen++'s manipulation capabilities through four tasks in both tabletop and shelf settings, as shown in Fig.~\ref{figure_manipulation}. The evaluation aims to validate the model's effectiveness in enabling feasible pick-and-place operations by integrating object and spatial affordance reasoning. Tasks 1 and Task 3 assess basic object affordance understanding by distinguishing object categories and spatial relations for localizing the leftmost fruit, the rightmost basket, and the orange that is farthest from the toy car. Task 2 evaluates object affordance prediction for the handle through mug grasping, coupled with spatial affordance to identify vacant space on the bottom shelf. Task 4 requires more complex affordance reasoning, which involves identifying a drawer handle for grasping to open it, discriminating between a real screwdriver and a toy one, and locating free space inside the drawer for placement. These results highlight RoboAfford-Qwen++'s advanced ability in object affordance prediction and spatial reasoning for real-world manipulation tasks.

The quantitative comparison in Tab.~\ref{table:realworld_manipulation} shows that RoboAfford-Qwen++ outperforms other models in real-world manipulation tasks, achieving an average success rate (SR) of 61.4\%, significantly surpassing GPT-4o~\cite{hurst2024gpt} (11.4\%), Qwen2.5-VL-7B~\cite{bai2025qwen2} (5.7\%), and RoboPoint~\cite{yuan2024robopoint} (35.7\%). RoboAfford-Qwen++ excels in both object affordance tasks (grasping an apple, mug, and opening a drawer) and free space tasks (placement in a bowl, on a plate, and in the cabinet). Specifically, it achieves high success rates in grasping an apple (7/10), a mug (8/10), and opening a drawer (4/10), demonstrating a robust understanding of object interactions. In free space tasks, RoboAfford-Qwen++ consistently performs well for placing objects in a bowl (8/10), on a plate (8/10), and in cabinet (6/10 on the top layer and 5/10 on the bottom layer), highlighting its advanced spatial reasoning capabilities. The 25.7\% improvement over the strongest baseline (RoboPoint~\cite{yuan2024robopoint}) emphasizes RoboAfford-Qwen++'s strengths in precise object manipulation and accurate spatial reasoning, validating its effectiveness for real-world robotic applications.

\begin{figure*}[!h]
\centering
  \includegraphics[width=0.90\textwidth]{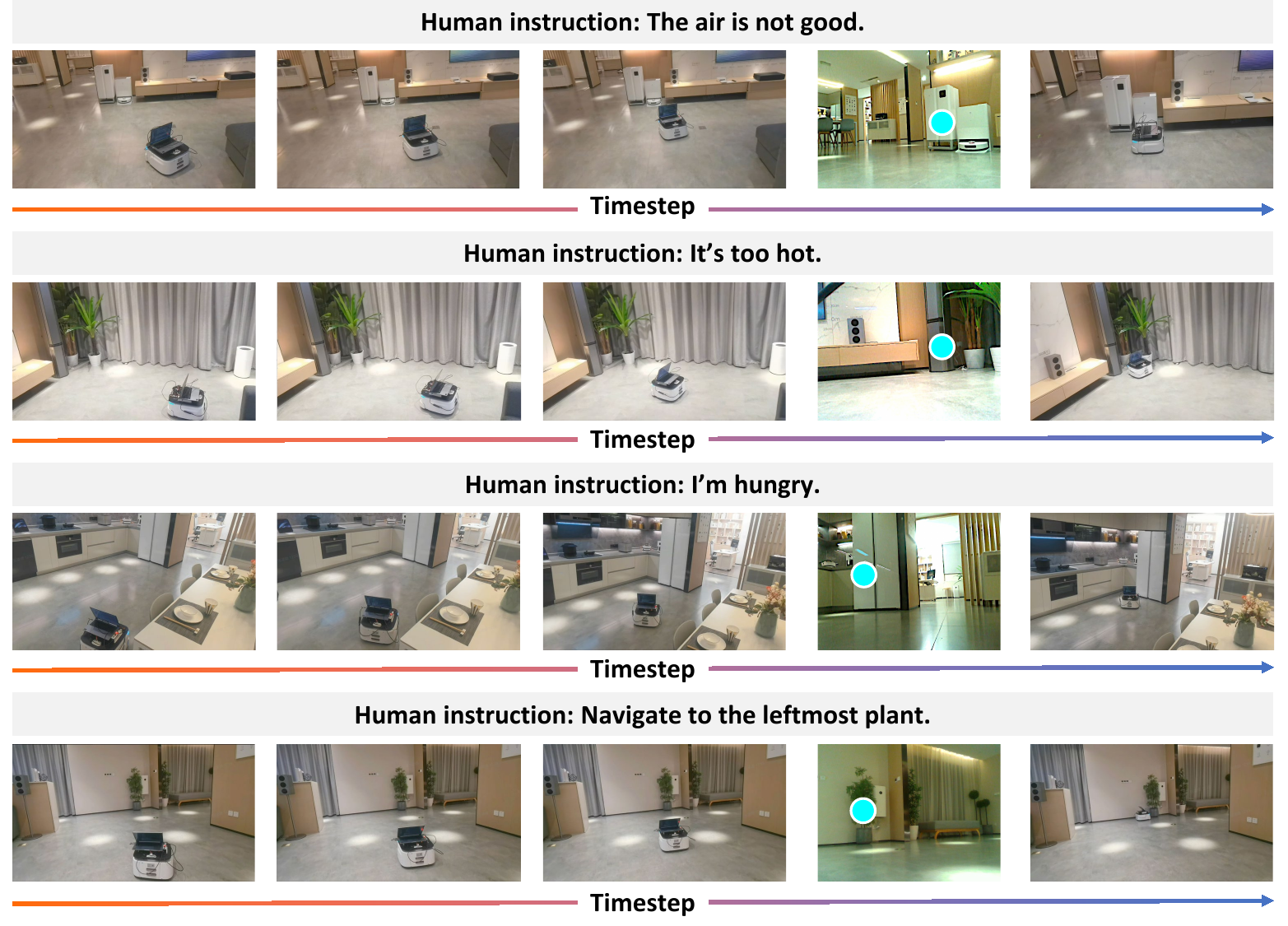}
  \vspace{-6pt}
  \caption{Results of deploying RoboAfford-Qwen++ model to downstream robotic navigation tasks.}
  \label{figure_navigation}
\end{figure*}

\begin{table*}[!h]
\centering
\caption{Real-world navigation comparison of different methods.}
\label{table:realworld_navigation}
\resizebox{0.96\textwidth}{!}{
\begin{tabular}{l|ccccccc|c}
\toprule
\textbf{Models} & 
\multicolumn{1}{c}{\textbf{Trash Can}} & 
\multicolumn{1}{c}{\textbf{Potted Plant}} & 
\multicolumn{1}{c}{\textbf{Air Purifier}} &
\multicolumn{1}{c}{\textbf{Fruit}} & 
\multicolumn{1}{c}{\textbf{Coffee Machine}} & 
\multicolumn{1}{c}{\textbf{Refrigerator}} &
\multicolumn{1}{c|}{\textbf{Free Space}} &
\multicolumn{1}{c}{\textbf{Avg. SR$\uparrow$}}\\
\midrule
GPT-4o~\cite{hurst2024gpt}  & 3/10 & 2/10 & 1/10 & 4/10 & 3/10 & 2/10 & 1/10 & 22.9\%\\
Qwen2.5-VL-7B~\cite{bai2025qwen2}   & 1/10 & 0/10 & 0/10 & 2/10 & 1/10 & 0/10 & 0/10 & 5.7\%\\
RoboPoint~\cite{yuan2024robopoint}   & 4/10 & 3/10 & 2/10 & 6/10 & 4/10 & 3/10 & 2/10 & 34.3\%\\
\midrule
\rowcolor{blue!5}
\textbf{RoboAfford-Qwen++ (Ours)}    & \textbf{7/10} & \textbf{6/10} & \textbf{6/10} & \textbf{9/10} & \textbf{8/10} & \textbf{7/10} & \textbf{6/10} & \textbf{70.0\% (+35.7\%$\uparrow$)}\\
\bottomrule
\end{tabular}
}
\vspace{-9pt}
\end{table*}

\textbf{Robotic Navigation} We further evaluate RoboAfford-Qwen++'s performance in downstream robotic navigation tasks through four household scenarios, as shown in Fig.~\ref{figure_navigation}. These tasks assess the model’s ability to translate affordance understanding into effective path planning and goal-oriented navigation. In Task 1, the model successfully localizes an air purifier and plans a feasible path using object affordance and real-time depth perception. Tasks 2 and 3 emphasize the model's proficiency in interpreting complex high-level instructions in living room and kitchen settings, with the model identifying the air conditioner and refrigerator based on functional attributes. Task 4 shows the model's capability in precise understanding of spatial relationships by navigating to the leftmost plant among multiple candidates.

Quantitative results in Tab.~\ref{table:realworld_navigation} demonstrate RoboAfford-Qwen++'s significant advantages over state-of-the-art models in real-world navigation tasks. It achieves a 70.0\% average success rate (SR), markedly higher than GPT-4o~\cite{hurst2024gpt} (22.9\%) and RoboPoint~\cite{yuan2024robopoint} (34.3\%).  Specially, it excels in fruit localization (9/10 success rate) and maintains strong performance on complex targets like refrigerators (7/10) and free space navigation (6/10). The model's 35.7\% improvement over the best baseline underscores its strengths in object affordance understanding and free space reasoning, validating it's effectiveness in real-world navigation.

\section{Conclusion}
In this paper, we introduce \textbf{\textit{RoboAfford++}}, a generative AI-enhanced dataset for object and spatial affordance learning, comprising 2.0 million question-answer pairs with fine-grained 2D point annotations. 
This dataset enhances VLMs in both robotic manipulation and navigation by improving their ability to reason about object affordance for functional grasping and spatial affordance for placement or movement.
We also present \textbf{\textit{RoboAfford-Eval}}, a benchmark with 338 manually annotated questions to evaluate VLMs on three core tasks: object affordance recognition, object affordance prediction, and spatial affordance localization. 
Extensive experiments reveal the limitations of current VLMs in fine-grained affordance reasoning, while models fine-tuned on RoboAfford++ demonstrate substantial improvements in predicting object and spatial affordances.
This work bridges the critical gap between high-level semantic understanding and low-level physical interactions in real-world robotic scenarios, advancing affordance-aware learning for both manipulation and navigation tasks. 
In the future, we will expand RoboAfford++ to encompass more diverse environments and complex tasks, further strengthening VLMs' spatial reasoning capabilities and supporting the development of more intelligent and adaptive robotic systems.

\addtolength{\textheight}{-12cm}   

\addtolength{\textheight}{12cm} 
\bibliographystyle{IEEEtran}
\balance
\bibliography{references}

\end{document}